\documentclass[conference]{IEEEtran}

\IEEEoverridecommandlockouts
\usepackage{cite}
\usepackage{amsmath,amssymb,amsfonts}
\usepackage{algorithmic}
\usepackage{graphicx}
\usepackage{textcomp}
\usepackage{booktabs}
\usepackage{xcolor}
\def\BibTeX{{\rm B\kern-.05em{\sc i\kern-.025em b}\kern-.08em
    T\kern-.1667em\lower.7ex\hbox{E}\kern-.125emX}}
\begin{document}

\title{Image-to-Image Translation with Diffusion Transformers and CLIP-Based Image Conditioning\\
}

\author{
\IEEEauthorblockN{1\textsuperscript{st} Qiang Zhu}
\IEEEauthorblockA{\textit{Department of Mechanical and Aerospace Engineering} \\
\textit{University of Houston} \\
Houston, TX, USA \\
qzhu11@uh.edu \\
ORCID: 0009-0002-0981-0635} 

\vspace{+1em}

\IEEEauthorblockN{3\textsuperscript{rd} Menghao Huo}
\IEEEauthorblockA{\textit{School of Engineering} \\
\textit{Santa Clara University} \\
Santa Clara, CA, USA \\
menghao.huo@alumni.scu.edu \\
ORCID: 0009-0000-0076-5343}

\and

\IEEEauthorblockN{2\textsuperscript{nd} Kuan Lu*}
\IEEEauthorblockA{\textit{School of Electrical and Computer Engineering} \\
\textit{Cornell University} \\
Ithaca, NY, USA \\
kl649@cornell.edu \\
ORCID: 0009-0003-5744-9247} 

\vspace{+1em}

\IEEEauthorblockN{4\textsuperscript{th} Yuxiao Li}
\IEEEauthorblockA{\textit{Department of Electrical and Computer Engineering} \\
\textit{Northeastern University} \\
Boston, MA, USA \\
li.yuxiao@northeastern.edu \\
ORCID: 0009-0002-9813-5925} 

\thanks{*Corresponding author: Kuan Lu (kl649@cornell.edu)}

}

\maketitle

\begin{abstract}
Image-to-image translation aims to learn a mapping between a source and a target domain, enabling tasks such as style transfer, appearance transformation, and domain adaptation. In this work, we explore a diffusion-based framework for image-to-image translation by adapting Diffusion Transformers (DiT), which combine the denoising capabilities of diffusion models with the global modeling power of transformers. To guide the translation process, we condition the model on image embeddings extracted from a pre-trained CLIP encoder, allowing for fine-grained and structurally consistent translations without relying on text or class labels. We incorporate both a CLIP similarity loss to enforce semantic consistency and an LPIPS perceptual loss to enhance visual fidelity during training. We validate our approach on two benchmark datasets: face2comics, which translates real human faces to comic-style illustrations, and edges2shoes, which translates edge maps to realistic shoe images. Experimental results demonstrate that DiT, combined with CLIP-based conditioning and perceptual similarity objectives, achieves high-quality, semantically faithful translations, offering a promising alternative to GAN-based models for paired image-to-image translation tasks.
\end{abstract}

\begin{IEEEkeywords}
computer vision, diffusion models, generative models, vision transformers, paired image-to-image translation, CLIP, semantic guidance
\end{IEEEkeywords}

\section{Introduction}
Paired Image-to-image translation is a fundamental computer vision task that aims to map images from one domain to another, often with the objective of altering style, structure, or modality while preserving semantic content. It has broad applications across domains such as medical imaging, autonomous driving, virtual try-on systems, and artistic style transfer. Generative Adversarial Networks (GANs) \cite{goodfellow2014generative} marked a significant milestone in generative modeling. GANs have significantly expanded the scope of applications, including generating images, translating between visual domains, and enhancing image resolution. In particular, Pix2Pix \cite{isola2017image} demonstrated the effectiveness of GANs in supervised image-to-image translation by learning a mapping between paired images using a conditional GAN framework. Following this, numerous GAN-based methods \cite{karras2019style,zhang2019self} have been developed and widely adopted across industries for various domain adaptation tasks. Despite their success, GANs come with several limitations. They often suffer from training instability, mode collapse, and challenges in generating high-resolution or detail-rich images. These issues become more pronounced in tasks requiring precise structural preservation and stylistic accuracy—such as facial stylization or domain-specific artistic rendering.

Recently, diffusion models\cite{ho2020denoising} have become a new choice for generative tasks by predicting the denoising process. While diffusion models offer superior image quality and training stability, adapting them to conditional image-to-image translation remains a growing research area. To further enhance computational efficiency, Latent Diffusion Models (LDMs) \cite{rombach2022high} transfer the denoising operation from high-dimensional pixel space to a more compact latent representation, significantly accelerating training while maintaining high-quality results. The rise of Transformers \cite{vaswani2017attention} has further revolutionized diffusion models. Peebles et al. \cite{peebles2023scalable} substituted the conventional U-Net generative backbone in diffusion frameworks with a transformer-driven backbone, known as diffusion transformer(DiT). By leveraging the self-attention mechanism of transformers, DiT achieves higher-quality, high-resolution image generation and excellent scalability.

In this work, we explore CLIP-based\cite{radford2021learning} image conditioning for DiT-based image-to-image translation. Unlike traditional conditional models that rely on text or class labels, our method leverages image embeddings extracted from a pretrained CLIP encoder to steer the diffusion process, enabling the model to better grasp subtle visual relationships between the source and target domains. We focus our study on the paired translation from real human faces to comic-style illustrations, a task that demands both semantic preservation and stylistic transformation. By conditioning the diffusion transformer on real face embeddings and incorporating perceptual and semantic similarity losses (CLIP and LPIPS\cite{zhang2018unreasonable}), we ensure structural consistency and stylized accuracy in the generated outputs. The main contributions of this study can be summarized as follows: 

\begin{itemize}
    \item We introduce a CLIP-conditioned DiT framework for image-to-image translation, using image embeddings rather than class labels.
    \item We introduce a loss formulation that incorporates both perceptual (LPIPS) and semantic (CLIP) consistency, improving output quality and style fidelity.
    \item We demonstrate the effectiveness of our approach on real2comic and edges2shoes paired datasets, achieving high-quality, identity-preserving translations compared to the traditional GAN-based method.
\end{itemize}

\section{Related Work}
\subsection{Denoising Diffusion Probabilistic Models (DDPMs)}

Denoising Diffusion Probabilistic Models (DDPMs) \cite{ho2020denoising} are generative models that synthesize data by reversing a noise corruption process. During the forward diffusion phase, the initial data point \(\mathbf{x}_0\) undergoes a progressive corruption process, resulting in a sequence of latent variables \(\mathbf{x}_1, \mathbf{x}_2, \ldots, \mathbf{x}_T\). This transformation follows the probabilistic transition:

\begin{equation}
q(\mathbf{x}_t \mid \mathbf{x}_{t-1}) = \mathcal{N}(\mathbf{x}_t; \sqrt{1 - \beta_t} \, \mathbf{x}_{t-1}, \beta_t \, \mathbf{I})
\end{equation}

Here, the scalar \(\beta_t\) determines the level of Gaussian noise introduced at each timestep. To recover the source input, the reverse process attempts to reconstruct \(\mathbf{x}_0\) from the noisy sample \(\mathbf{x}_T\), using a neural network parameterized distribution:

\begin{equation}
p_{\theta}(\mathbf{x}_{t-1} \mid \mathbf{x}_t) = \mathcal{N}(\mathbf{x}_{t-1}; \mu_{\theta}(\mathbf{x}_t, t), \Sigma_{\theta}(\mathbf{x}_t, t))
\end{equation}

DDPMs provide several key advantages over GANs. Their iterative denoising process avoids mode collapse and stabilizes training, leading to diverse and high-quality image generation. However, the computational cost of the reverse process, which typically requires thousands of denoising steps, remains a major limitation. 

\subsection{Latent Diffusion Models (LDMs)}  

LDMs \cite{rombach2022high} are an efficient extension of DDPMs designed to reduce the computational cost of diffusion-based image generation. Instead of operating directly on high-dimensional pixel space, LDMs conduct the denoising process in a latent space with a lower dimension. The core idea is to first encode the input image \(\mathbf{x}_0\) into a compact latent representation \(\mathbf{z}_0\) using a pre-trained encoder \(E(\cdot)\):
\begin{equation}
    \mathbf{z}_0 = E(\mathbf{x}_0)
\end{equation}

The forward diffusion process then corrupts \(\mathbf{z}_0\) into a series of noisy latent variables \(\mathbf{z}_t\) using the same transition as in DDPMs, the reverse process aims to rebuild the original latent \(\mathbf{z}_0\) from the noised variable \(\mathbf{z}_T\) through a parameterized distribution. After obtaining $z_0$, the decoder $D(\cdot)$ reconstructs it into the image domain, yielding the final result $\hat{x}_0$:
\begin{equation}
    \hat{\mathbf{x}}_0 = D(\mathbf{z}_0)
\end{equation}

LDMs achieve substantial computational efficiency since the diffusion process is applied to a much smaller latent space rather than the pixel space. This reduction in dimensionality significantly decreases the memory and time required for training, allowing LDMs to scale to high-resolution image generation. The approach has been widely adopted in image synthesis, text-to-image generation, and other generative tasks.

\subsection{Transformers and Vision Transformers (ViT)}  

Transformers \cite{vaswani2017attention} were originally developed for natural language processing (NLP) and other LLM time-series tasks\cite{xu2025drift2matrix, behari2024decision, liu2024rag,zhang2025ratt}. The key component of transformers is the self-attention mechanism, which can effciently capture long-range dependencies in the input data. Unlike traditional convolutional neural networks (CNNs) \cite{lu2023deep} that operate on local receptive fields, transformers process the entire input as a sequence, enabling global context modeling\cite{tao2024nevlp,ying2024feature,glenn2024blendsql}. 

The Vision Transformer (ViT) \cite{dosovitskiy2020image} introduced the idea of applying transformer architecture directly to image data. Instead of using convolutional layers, ViT divides an input image \(\mathbf{x} \in \mathbb{R}^{H \times W \times C}\) into non-overlapping patches of size \(P \times P\). Then each patch will be flattened into a vector and projected into an embedding space, forming a sequence of tokens \(\mathbf{z}_p \in \mathbb{R}^{N \times D}\), where \(N = \frac{H \cdot W}{P^2}\) is the total number of patches. The process can be formalized as:
\begin{equation}
    \mathbf{z}_p = [\mathbf{x}_1 \mathbf{P}, \mathbf{x}_2 \mathbf{P}, \dots, \mathbf{x}_N \mathbf{P}] + \mathbf{E}_{\text{pos}}
\end{equation}
where \(\mathbf{P}\) is the learnable projection matrix, and \(\mathbf{E}_{\text{pos}}\) represents the position embeddings that retain spatial information. The resulting sequence of tokens is passed through multiple layers of multi-head self-attention and feed-forward neural networks. The final classification token is used for downstream tasks such as classification, segmentation, or image generation. 

A key benefit of transformers over CNNs is their ability to model long-range dependencies and global context. This is particularly useful for tasks like image generation and image-to-image translation, where understanding global structure is crucial. However, the quadratic complexity of self-attention with respect to sequence length poses a computational challenge for large images. Recent works have introduced improvements like Swin Transformers \cite{liu2021swin}, which reduce the computational burden using window-based attention, and Pyramid Vision Transformers (PVT) \cite{wang2021pyramid}, which enable hierarchical feature extraction similar to CNNs.

\subsection{Diffusion Transformer (DiT)}  

Diffusion Transformer (DiT)  integrates the architecture of transformers with latent diffusion models, replacing the conventional U-Net backbone used in LDMs. Unlike U-Net and its variants\cite{li2024ucloudnet,dan2024image,li2025ddunet}, which relies on convolutional layers, DiT employs a ViT-style backbone, allowing it to capture long-range dependencies and global contextual relationships. This design improves scalability, enabling high-resolution image generation with greater computational efficiency.

The forward diffusion process adds noise to the latent representation \(\mathbf{z}_0\), following the same transition as in DDPMs, while the reverse diffusion process denoises the noisy latent \(\mathbf{z}_t\) back to \(\mathbf{z}_0\) through a parameterized distribution. Unlike U-Net-based models that operate at multiple resolutions, the transformer backbone processes a flat token sequence, handling all image tokens simultaneously. Conditions, such as class labels or conditioning images, can be incorporated during training, guiding the model to generate class-specific outputs. By leveraging self-attention, DiT models global spatial relationships, making it well-suited for high-resolution image synthesis. Its transformer-based backbone also provides better scalability and robustness, especially for large, high-dimensional datasets.

\section{Method}
\subsection{Data Pre-processing}

The data preprocessing pipeline used in this study is illustrated on the left side of Fig.~\ref{fig:method}. For images from the target domain, we first resize them to \(256 \times 256\) to match the input size expected by the pre-trained Variational Autoencoder (VAE)\cite{kingma2013auto}. These images are then encoded by the VAE into a lower-dimensional latent space, reducing computational cost while preserving important structural information. Following encoding, a standard ViT patchification process is applied: the latent feature maps are segmented into distinct, non-overlapping patches, with each patch subsequently transformed via a linear projection into a hidden-dimensional embedding space.

Conditioning images, which provide guidance to the model during generation, are preprocessed separately. They are resized to \(224 \times 224\) to align with the input size requirements of the pre-trained CLIP-ViT-L/14 model \cite{radford2021learning}. The CLIP encoder then extracts semantic latent representations from these images. These CLIP embeddings are projected into the same hidden space and summed with the timestep embeddings, enabling conditional guidance throughout the diffusion process.
\subsection{Architecture}

\begin{figure*}
    \centering
    \includegraphics[width=\linewidth]{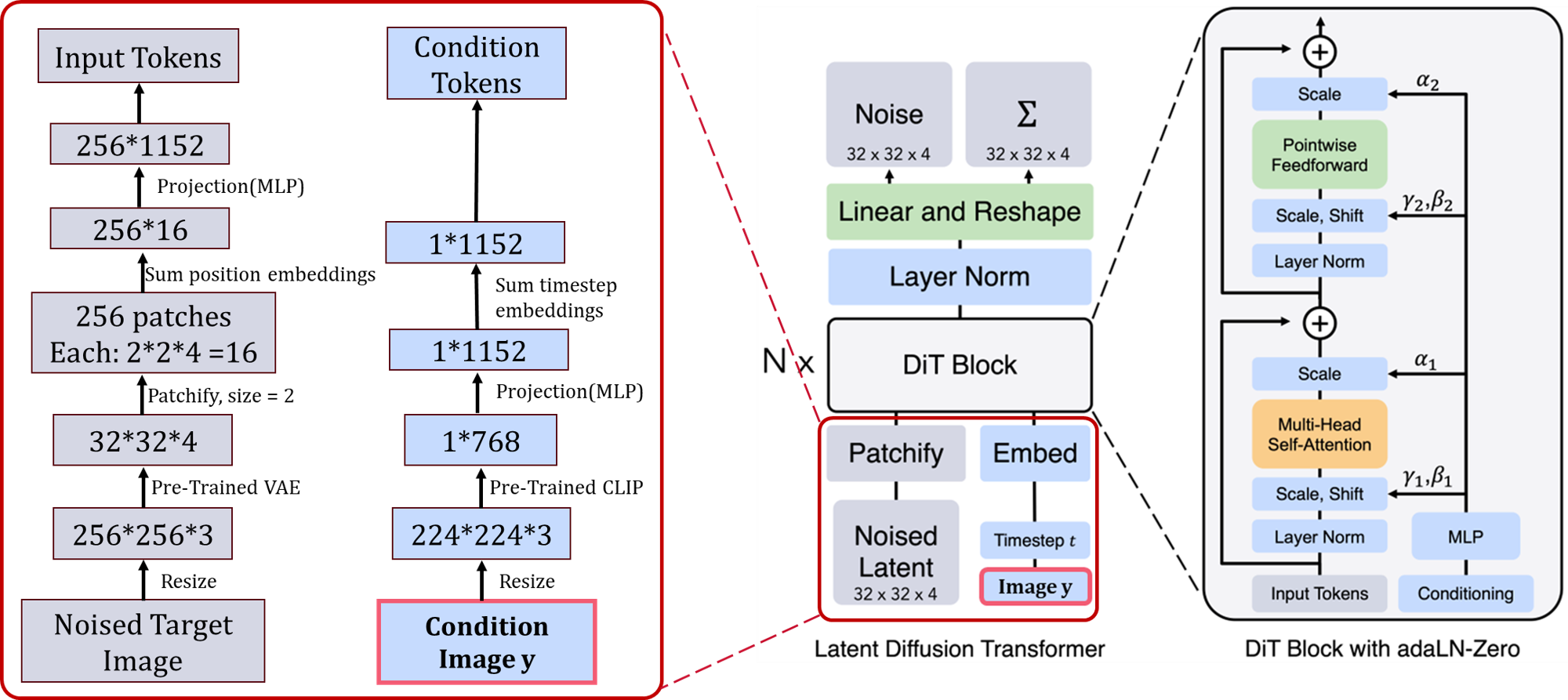}
    \caption{Image-Conditioned Diffusion Transformer Architecture}
    \label{fig:method}
\end{figure*}

The proposed method builds on the Diffusion Transformer (DiT) to enable paired image-to-image translation with conditional guidance using latent representations extracted from input images. Unlike the original DiT, which conditions the denoising process on class labels, our method uses image embeddings from a pre-trained CLIP encoder to provide fine-grained semantic conditioning. This enables the model to better preserve structural information and stylistic consistency during translation. The overall architecture is illustrated in Fig.~\ref{fig:method}.

The process begins by encoding target domain images using a pre-trained VAE into a lower-dimensional latent space. These VAE latents serve as the inputs to the diffusion model. Gaussian noise is then added to the latents according to a predefined noise schedule, following the standard denoising diffusion probabilistic model (DDPM) training protocol.

The noisy latent representations are divided into non-overlapping patches, flattened, and projected into patch embeddings. Positional embeddings are added to preserve spatial information, following the patchification strategy of Vision Transformers (ViTs) \cite{dosovitskiy2020image}. These patch embeddings are then processed through the DiT blocks, which apply sequences of multi-head self-attention (MSA) and feed-forward networks (FFN) to learn global dependencies across patches.

To enable image-conditioned translation, we extract semantic features from the source images using a pre-trained CLIP-ViT-L/14 model. The resulting CLIP embeddings are projected into the hidden dimension and injected into the diffusion process through a combination of multi-head cross-attention and adaptive normalization using AdaLN-Zero blocks.

In particular, the DiT blocks incorporate conditional information through two key mechanisms:
\begin{itemize}
    \item \textbf{Cross-Attention:} Patch embeddings from the noisy latent attend to the CLIP embeddings of the source image, allowing the model to query domain-specific features and maintain semantic alignment.
    \item \textbf{AdaLN-Zero Modulation:} Feature activations are adaptively modulated using the conditioning inputs. The AdaLN-Zero block introduces scaling parameters (\(\alpha_1, \alpha_2\)) for residual connections and gain/bias parameters (\(\gamma_1, \beta_1, \gamma_2, \beta_2\)) for layer normalization, enabling fine-grained control over the denoising dynamics.
\end{itemize}

After the DiT backbone, the model predicts the added noise for each patch embedding, conditioned on both the timestep embeddings and the CLIP-derived image features. Through iterative refinement, the model progressively denoises the latent representations. Finally, the cleaned latent is decoded back into pixel space using the pre-trained VAE decoder, resulting in an output image that reflects the target domain style while preserving the structural content of the input. By operating in latent space and using strong semantic conditioning from real images, our framework enables efficient, scalable, and high-fidelity paired image-to-image translation.

\subsection{Loss Function Design}
We design a composite loss function that balances pixel-level accuracy, perceptual fidelity, and semantic consistency. The overall loss is formulated as:
\begin{equation}
\mathcal{L}_{\text{total}} = \lambda_{\text{rec}} \cdot \mathcal{L}_{\text{rec}} + \lambda_{\text{LPIPS}} \cdot \mathcal{L}_{\text{LPIPS}} + \lambda_{\text{CLIP}} \cdot \mathcal{L}_{\text{CLIP}},
\end{equation}

\noindent where $\lambda_{\text{rec}}$, $\lambda_{\text{LPIPS}}$, and $\lambda_{\text{CLIP}}$ are scalar weights controlling the influence of each loss term.

\paragraph{Reconstruction Loss.} We employ an L1 reconstruction loss to enable the generated image to resemble the ground truth target image at a low level, which can be represented as:
\begin{equation}
\mathcal{L}_{\text{rec}} = \| \hat{x}_{\text{target}} - x_{\text{target}} \|_1,
\end{equation}

\noindent where $x_{\text{target}}$ is the ground truth image in the target domain, and $\hat{x}_{\text{target}}$ is the generated image.

\paragraph{Perceptual Loss (LPIPS).} To enforce perceptual similarity and retain high-level structure, we use the LPIPS loss \cite{zhang2018unreasonable}, which compares deep features extracted from a pre-trained network:
\begin{equation}
\mathcal{L}_{\text{LPIPS}} = \| \phi_{\text{LPIPS}}(\hat{x}_{\text{target}}) - \phi_{\text{LPIPS}}(x_{\text{target}}) \|_2^2,
\end{equation}

\noindent where $\phi_{\text{LPIPS}}(\cdot)$ denotes the LPIPS feature extractor.

\paragraph{Semantic Consistency Loss (CLIP).} To ensure that the translation process aligns semantically with the input image from the source domain, we incorporate a CLIP-based similarity loss \cite{radford2021learning}. We compute the cosine similarity between the generated image and the conditioning image, both encoded using a pre-trained CLIP image encoder:
\begin{equation}
\mathcal{L}_{\text{CLIP}} = 1 - \cos \left( \phi_{\text{CLIP}}(\hat{x}_{\text{target}}), \phi_{\text{CLIP}}(x_{\text{source}}) \right),
\end{equation}

\noindent where $x_{\text{source}}$ is the input image from the source domain and $\phi_{\text{CLIP}}(\cdot)$ is the CLIP image encoder.

\medskip

This combination of losses allows the model to generate visually convincing target domain images while preserving the structural and semantic essence of the source domain inputs.

\section{Experiments}

\subsection{Dataset}

To demonstrate the feasibility of the proposed method, we trained the model on two paired image dataset: \textit{face2comics}  \cite{Spirin2021face2comics} and \textit{edges2shoes}\cite{fine-grained, xie15hed}. The face2comics dataset consists of paired real human face images and corresponding comic-style illustrations, providing a benchmark for evaluating stylization while preserving identity. For face2comics, we use 9,000 pairs of images for training and the rest 1,000 pairs for validation and testing. For edges2shoes, it contains pairs of shoe edge maps and corresponding realistic shoe images, enabling evaluation of structural-to-appearance translation tasks. For edges2shoes, 49,825 image pairs are used for training and 200 pairs for testing. Both datasets offer paired supervision, allowing us to directly train the model to translate between the source and target domains with strong ground truth alignment.

\subsection{Hyperparameters}

We trained our model based on a pre-trained DiT, using the conditioning input extracted from the source image. The model parameters and architecture settings are summarized in Table~\ref{tab:parameter}, and the training hyperparameters are listed in Table~\ref{tab:hyperparameters}. Training was performed on Google Colab using an NVIDIA A100 GPU, achieving an average training speed of 0.45 training steps per second.

\begin{table}[!h]
\centering
\caption{Parameters of pre-trained models.}
\label{tab:parameter}
\begin{tabular}{lccc}
\hline
\textbf{Pre-trained Model} & \textbf{Layers $N$} & \textbf{Hidden Size $d$} & \textbf{Heads} \\
\hline
DiT-XL-256 \cite{peebles2023scalable} & 28 & 1156 & 16 \\
\hline
\end{tabular}
\end{table}

\begin{table}[!h]
\centering
\caption{Hyperparameter settings.}
\label{tab:hyperparameters}
\begin{tabular}{l c}
\hline
\textbf{Hyperparameter}    & \textbf{Value} \\
\hline
Learning rate              & 0.0001         \\
Weight decay               & 0.0001         \\
Optimizer                  & AdamW          \\
Number of iterations       & 40,000         \\
Batch size                 & 64             \\
\hline
\end{tabular}
\end{table}

\subsection{Training Results}
\begin{figure}[h]
    \centering
    \includegraphics[width=\linewidth]{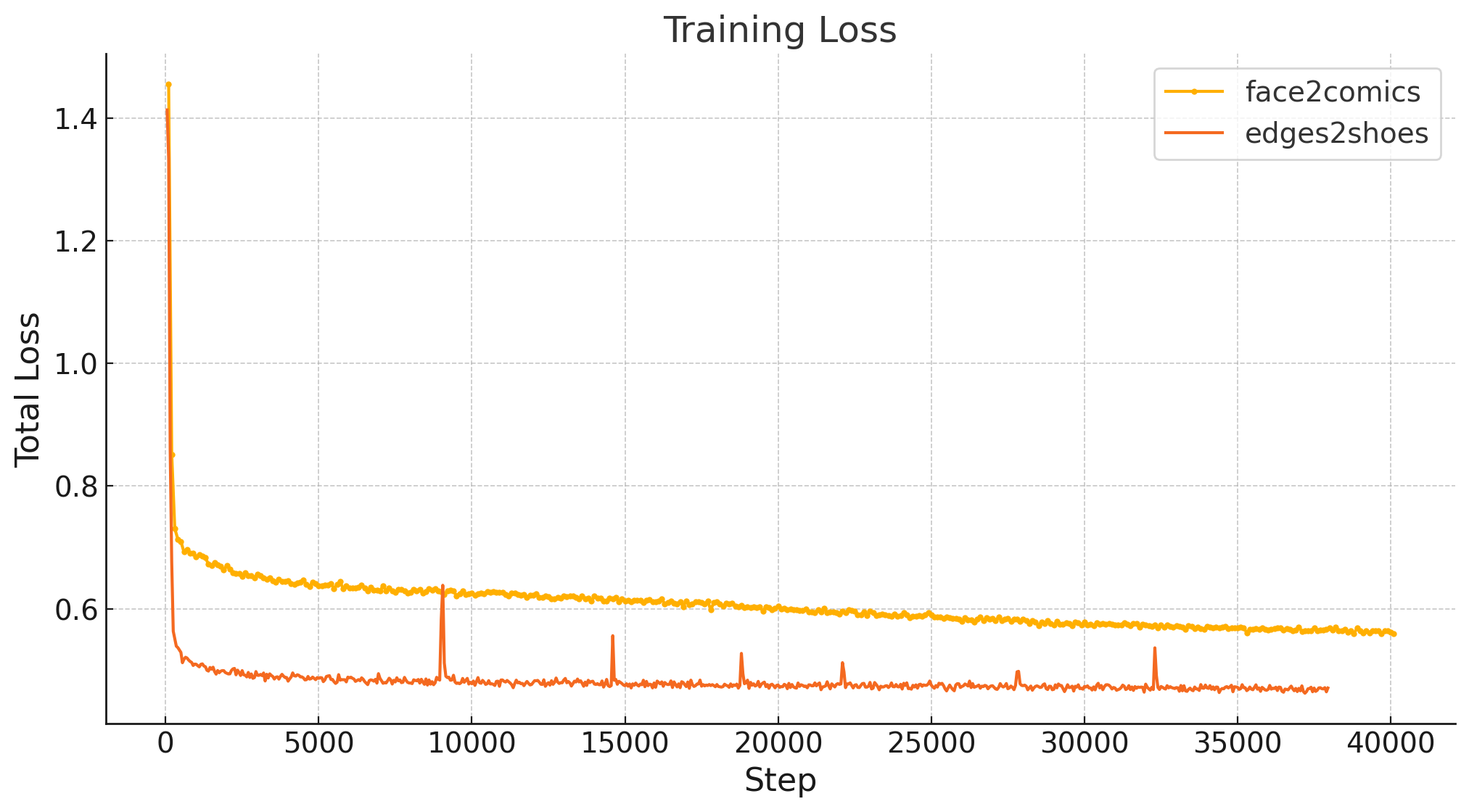}
    \caption{The training loss over iteration steps}
    \label{fig:LOSS}
\end{figure}

The training loss curves shown in Fig.~\ref{fig:LOSS} exhibit a gradual decrease over time, indicating that the model is effectively learning from the data. Although the raw loss trajectories display minor fluctuations due to the stochastic nature of the training process, the overall trend reveals consistent convergence. The loss curves provide a clear view of the general learning behavior, highlighting a steady reduction in loss values as training progresses. Additionally, we observe that the model trained on the larger edges2shoes dataset converges to a lower final loss compared to the face2comics dataset, suggesting that larger datasets contribute to better model optimization and generalization.

\subsection{Inference and Comparison}

\begin{figure}
    \centering
    \includegraphics[width=0.95\linewidth]{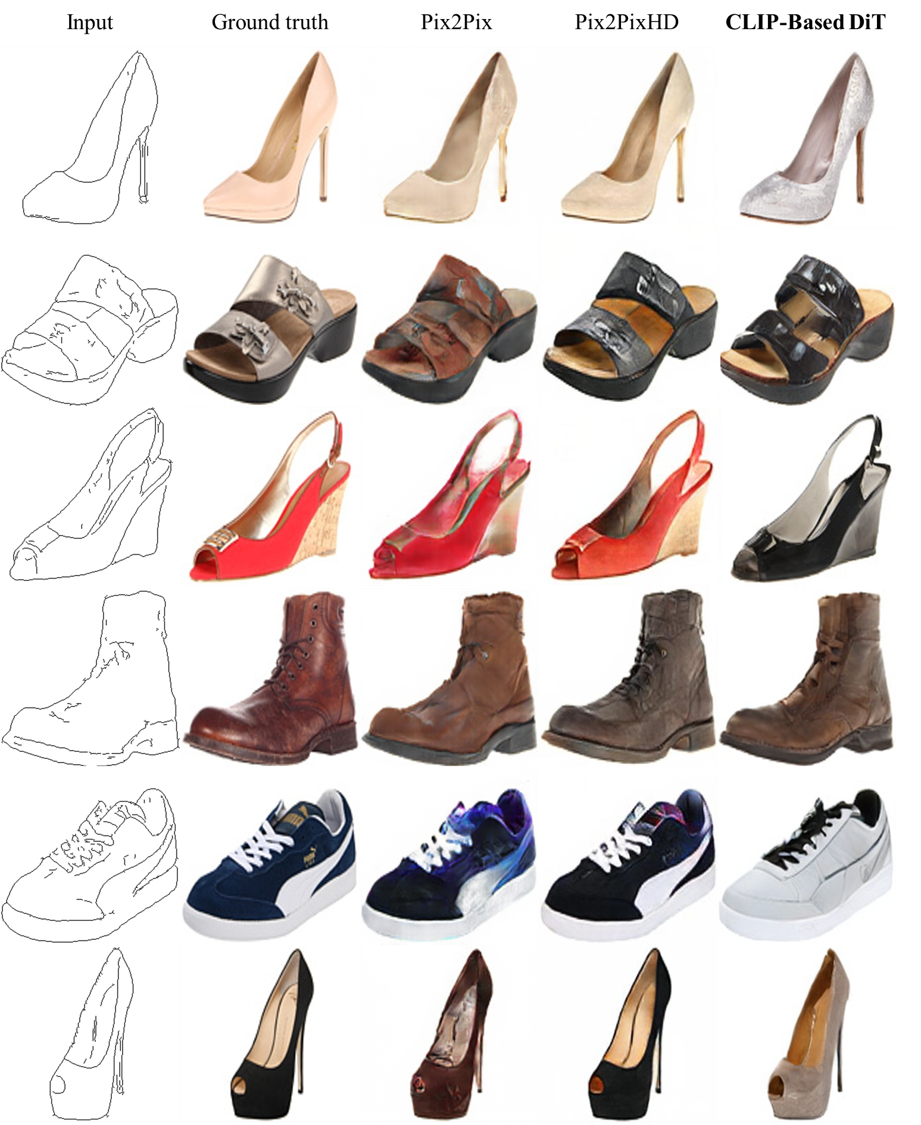}
\caption{Qualitative examples from the unseen edges2shoes dataset. Our method produces sharper, more detailed, and more realistic images compared to Pix2Pix and Pix2PixHD.}
    \label{fig:reslutt_shoes}
\end{figure}

\begin{figure}
    \centering
    \includegraphics[width=0.95\linewidth]{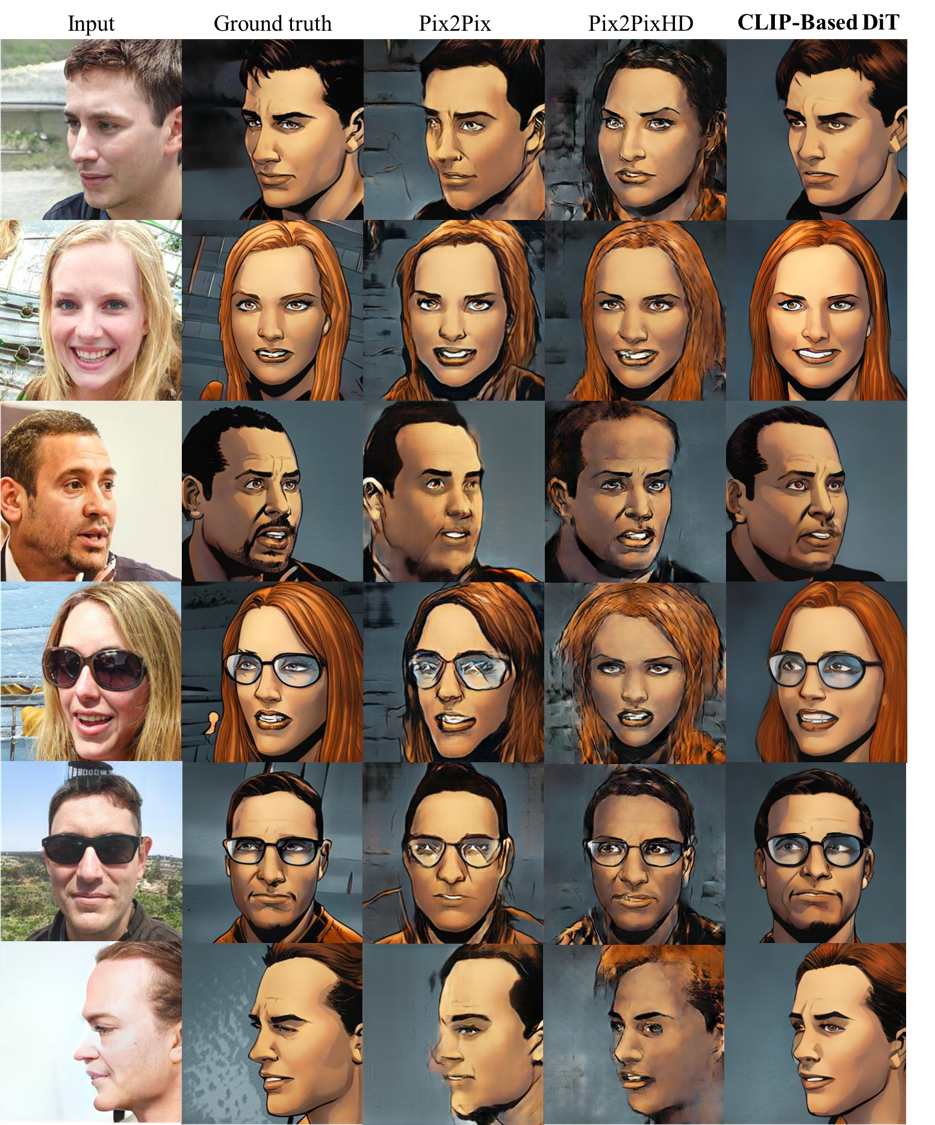}
\caption{Qualitative examples from the unseen face2comics dataset.The CLIP-conditioned DiT better preserves facial identity and stylization details compared to Pix2Pix and Pix2PixHD.}
    \label{fig:reslutt_faces}
\end{figure}

The inference process begins by inputting an unseen source image, which is perturbed by adding noise at a random timestep. The model then performs a denoising process to progressively refine the noisy input into a realistic target domain image. As shown in Fig.~\ref{fig:reslutt_shoes} and Fig.~\ref{fig:reslutt_faces}, we compare our method against two baselines: Pix2Pix and Pix2PixHD\cite{wang2018pix2pixHD}, using a pre-trained Pix2Pix model for edges2shoes and a newly trained Pix2Pix model for face2comics, both dates are newly trained with Pix2PixHD.

Visual comparisons reveal that our CLIP-conditioned DiT model generates images with noticeably higher quality, sharper details, and fewer artifacts. Unlike Pix2Pix, which often produces blurred regions and occasional mis-generation, the DiT outputs are sharp and consistent. In particular, our method better preserves fine details such as edges, highlights, textures, hairstyles, and glasses. The overall appearance of the generated images is smoother, more natural, and stylistically coherent.

Furthermore, we observe that the model achieves better results on the edges2shoes dataset compared to face2comics, especially in terms of preserving identity and fine-grained structural details. To evaluate the model's robustness on smaller datasets, we also trained it on the CMP Facade Database \cite{Tylecek13}, which contains only 400 training images. On this smaller dataset, Pix2Pix outperformed our DiT model, suggesting that diffusion transformers benefit significantly from larger training datasets. This observation aligns with the scaling properties of DiTs, which achieve stronger performance as dataset size and model capacity increase.

\subsection{Computational Cost}

\begin{table}[htbp]
\centering
\caption{Computation cost comparison}
\begin{tabular}{l|c|c|c}
\toprule
\textbf{Property} & \textbf{Pix2Pix} & \textbf{Pix2PixHD} & \textbf{Ours (DiT)} \\
\midrule
Batch Size              & 8            & 64            & 64 \\
Training Speed (steps/s)& 180          & 33.3          & 0.48 \\
Training Time           & 1.4 h        & 3.7 h         & 23.1 h \\
Epochs                  & 100          & 50            & 50 \\
Inference Speed (img/s) & 25.0 (0.04 s)& 1.2 (0.8 s)& 0.028 (36.1 s) \\
Model Size              & 207.6 MB     & 696 MB        & 12.92 GB \\
\bottomrule
\end{tabular}
\label{tab:model_comparison_clean}
\end{table}

All models were trained and evaluated using an NVIDIA A100 GPU under comparable settings. As shown in Table~\ref{tab:model_comparison_clean}, our DiT-based model incurs significantly higher computational cost compared to the GAN-based baselines. In particular, both training and inference times are much longer due to the iterative nature of the diffusion process and the complexity of the transformer backbone. Despite the heavier cost, DiT offers major benefits in terms of generation quality, structure preservation, and style consistency. Additionally, it avoids common GAN training issues such as mode collapse, resulting in more stable and robust convergence.

\section{Conclusion}
This work introduces an innovative framework for paired image-to-image translation by adapting the DiT architecture to incorporate image-based conditioning. Instead of relying on traditional class label embeddings, our approach leverages latent representations extracted from real images using a pre-trained CLIP encoder. The conditioning information is injected into the diffusion process through multi-head cross-attention and AdaLN-Zero blocks, allowing the model to capture domain-specific features and guide generation in a semantically meaningful way.

We validated our method on face2comics and edges2shoes paired datasets, demonstrating that CLIP-based image conditioning significantly improves translation quality, structural preservation, and stylistic consistency. Furthermore, our experiments show that DiT achieves substantially better generative performance compared to GAN-based baselines, especially on larger datasets such as edges2shoes, highlighting the scalability and stability advantages of diffusion-based models over adversarial training. While our approach introduces significantly higher computational cost in both training and inference—largely due to the iterative nature of diffusion sampling and the transformer-based architecture, we show that the quality improvements justify the trade-off. In particular, DiT requires longer training schedules and has slower inference speed compared to GAN models, but offers superior fidelity and robustness. In the future, we will try to extend our framework to the unpaired image-to-image translation setting by developing a Cycle-DiT model, incorporating cycle-consistency loss to enforce structural preservation without paired supervision. Additionally, longer training schedules and larger batch sizes will be explored to further improve output fidelity and convergence.

\bibliographystyle{IEEEtran}
\bibliography{egbib}

\end{document}